\documentclass[11pt,a4paper]{article}
\usepackage[hyperref]{acl2017}
\usepackage{times}
\usepackage{latexsym}

\usepackage{framed}
\usepackage{scrextend}
\usepackage{soul}
\usepackage{color}
\usepackage{booktabs}
\usepackage{paralist}
\usepackage{url}

\aclfinalcopy

\usepackage{enumerate}

\title{SemEval-2017 Task 8: RumourEval: Determining rumour veracity and support for rumours}

\author{Leon Derczynski$^{\heartsuit\ast}$ \and Kalina Bontcheva$^\heartsuit$ \and Maria Liakata$^\clubsuit$\\ \and \textbf{Rob Procter}$^\clubsuit$ \and \textbf{Geraldine Wong Sak Hoi}$^\diamondsuit$ \and \textbf{Arkaitz Zubiaga}$^\clubsuit$\\
\\
$\heartsuit$: Department of Computer Science, University of Sheffield, S1 4DP, UK\\
$\clubsuit$: Department of Computer Science, University of Warwick, CV4 7AL, UK\\
$\diamondsuit$:  swissinfo.ch, Bern, Switzerland\\
$\ast$: {\tt \small leon.d@shef.ac.uk}
}

\date{}

\begin{document}

\maketitle

\begin{abstract}
Media is full of false claims. Even Oxford Dictionaries named ``post-truth'' as the word of 2016. This makes it more important than ever to build systems that can identify the veracity of a story, and the nature of the discourse around it. RumourEval is a SemEval shared task that aims to identify and handle rumours and reactions to them, in text. We present an annotation scheme, a large dataset covering multiple topics -- each having their own families of claims and replies --  and use these to pose 
two concrete challenges as well as the results achieved by participants on these challenges.
\end{abstract}

\section{Introduction and Motivation}

Rumours are rife on the web. 
False claims affect people's perceptions of events and their behaviour, sometimes in harmful ways.
With the increasing reliance on the Web -- social media, in particular -- as a source of information and news updates by individuals, news professionals, and automated systems, the potential disruptive impact of rumours is further accentuated. 

The task of analysing and determining veracity of social media content has been of recent interest to the field of natural language processing.
After initial work~\cite{qazvinian2011rumor}, increasingly advanced systems and annotation schemas have been developed to support the analysis of rumour and misinformation in text~\cite{kumar2014detecting,zhang2015automatic,shao2016hoaxy,zubiaga2016analysing}.
Veracity judgment can be decomposed intuitively in terms of a comparison between assertions made in -- and entailments from -- a candidate text, and external world knowledge.
Intermediate linguistic cues have also been shown to play a role. 
Critically, based on recent work the task appears deeply nuanced and very challenging, while having important applications in, for example, journalism and disaster mitigation \cite{hermida2012tweets,procter2013police,veil2011work}.

We propose a shared task where participants analyse rumours in the form of claims made in user-generated content, and where users respond to one another within conversations attempting to resolve the veracity of the rumour. 
We define a rumour as a ``circulating story of questionable veracity, which is apparently credible but hard to verify, and produces sufficient scepticism and/or anxiety so as to motivate finding out the actual truth'' \cite{zubiaga2015towards}.
While breaking news unfold, gathering opinions and evidence from as many sources as possible as communities react becomes crucial to determine the veracity of rumours and consequently reduce the impact of the spread of misinformation.

Within this scenario where one needs to listen to, and assess the testimony of, different sources to make a final decision with respect to a rumour's veracity, we ran a task in SemEval consisting of two subtasks: (a) stance classification towards rumours, and (b) veracity classification. 
Subtask A corresponds to the core problem in crowd response analysis when using discourse around claims to verify or disprove them.
Subtask B corresponds to the AI-hard task of assessing directly whether or not a claim is false.


\begin{figure*}
\small
 \begin{framed}
  \textbf{SDQC support classification. Example 1:} \vspace{1em} \\
  \noindent \textbf{u1:} We understand there are two gunmen and up to a dozen hostages inside the cafe under siege at Sydney.. ISIS flags remain on display \#7News \textbf{[support]}
  \begin{addmargin}[2em]{0pt}
   \textbf{u2:} @u1 not ISIS flags \textbf{[deny]}
  \end{addmargin}
  \begin{addmargin}[2em]{0pt}
   \textbf{u3:} @u1 sorry - how do you know it's an ISIS flag? Can you actually confirm that? \textbf{[query]}
  \end{addmargin}
  \begin{addmargin}[4em]{0pt}
   \textbf{u4:} @u3 no she can't cos it's actually not \textbf{[deny]}
  \end{addmargin}
  \begin{addmargin}[2em]{0pt}
   \textbf{u5:} @u1 More on situation at Martin Place in Sydney, AU --LINK-- \textbf{[comment]}
  \end{addmargin}
  \begin{addmargin}[2em]{0pt}
   \textbf{u6:} @u1 Have you actually confirmed its an ISIS flag or are you talking shit \textbf{[query]}
  \end{addmargin}
   \vspace{1em}
  \textbf{SDQC support classification. Example 2:} \vspace{1em} \\
  \noindent \textbf{u1:} These are not timid colours; soldiers back guarding Tomb of Unknown Soldier after today's shooting \#StandforCanada --PICTURE-- \textbf{[support]}
  \begin{addmargin}[2em]{0pt}
   \textbf{u2:} @u1 Apparently a hoax. Best to take Tweet down. \textbf{[deny]}
  \end{addmargin}
  \begin{addmargin}[2em]{0pt}
   \textbf{u3:} @u1 This photo was taken this morning, before the shooting. \textbf{[deny]}
  \end{addmargin}
  \begin{addmargin}[2em]{0pt}
   \textbf{u4:} @u1 I don't believe there are soldiers guarding this area right now. \textbf{[deny]}
  \end{addmargin}
  \begin{addmargin}[4em]{0pt}
   \textbf{u5:} @u4 wondered as well. I've reached out to someone who would know just to confirm that. Hopefully get response soon. \textbf{[comment]}
  \end{addmargin}
  \begin{addmargin}[6em]{0pt}
   \textbf{u4:} @u5 ok, thanks. \textbf{[comment]}
  \end{addmargin}
 \end{framed}
 \caption{Examples of tree-structured threads discussing the veracity of a rumour, where the label associated with each tweet is the target of the SDQC support classification task.}
 \label{fig:example}
\end{figure*}

\subsection{Subtask A - SDQC Support/ Rumour stance classification}

Related to the objective of predicting a rumour's veracity, Subtask~A deals with the complementary objective of tracking how other sources orient to the accuracy of the rumourous story. 
A key step in the analysis of the surrounding discourse is to determine how other users in social media regard the rumour~\cite{procter2013reading}. 
We propose to tackle this analysis by looking at the conversation stemming from direct and nested replies to the tweet originating the rumour (source tweet). 

To this effect RumourEval provided participants with a tree-structured conversation formed of tweets replying to the originating rumourous tweet, directly or indirectly. Each tweet presents its own type of support with respect to the rumour (see Figure \ref{fig:example}). 
We frame this in terms of supporting, denying, querying or commenting on (SDQC) the original rumour~\cite{zubiaga2016analysing}.
Therefore, we introduce a subtask where the goal is to label the type of interaction between a given statement (rumourous tweet) and a reply tweet (the latter can be either direct or nested replies).

We note that superficially this subtask may bear similarity to SemEval-2016 Task 6 on stance detection from tweets~\cite{saif2016stance}, where participants are asked to determine whether a tweet is in favour, against or neither, of a given target entity (e.g. Hillary Clinton) or topic (e.g. climate change). 
Our SQDC subtask differs in two aspects. 
Firstly, participants needed to determine the objective support towards a rumour, an entire statement, rather than individual target concepts. 
Moreover, they are asked to determine additional response types to the rumourous tweet that are relevant to the discourse, such as a request for more information (questioning, Q) and making a comment (C), where the latter doesn't directly address support or denial towards the rumour, but provides an indication of the conversational context surrounding rumours. 
For example, certain patterns of comments and questions can be indicative of false rumours and others indicative of rumours that turn out to be true. 

Secondly, participants need to determine the type of response towards a rumourous tweet from a tree-structured conversation, where each tweet is not necessarily sufficiently descriptive on its own, but needs to be viewed in the context of an aggregate discussion consisting of tweets preceding it in the thread. 
This is more closely aligned with stance classification as defined in other domains, such as public debates \cite{Anand:2011:CRD:2107653.2107654}. 
The latter also relates somewhat to the SemEval-2015 Task 3 on Answer Selection in Community Question Answering \cite{alessandromoschitti2015semeval}, where the task was to determine the quality of responses in tree-structured threads in CQA platforms. 
Responses to questions are classified as `good', `potential' or `bad'. 
Both tasks are related to textual entailment and textual similarity. 
However, Semeval-2015 Task3 is clearly a question answering task, the platform itself supporting a QA format in contrast with the more free-form format of conversations in Twitter. 
Moreover, as a question answering task Semeval-2015 Task 3 is more concerned with relevance and retrieval whereas the task we propose here is about whether support or denial can be inferred towards the original statement (source tweet) from the reply tweets.

Each tweet in the tree-structured thread is categorised into one of the following four categories, following~\newcite{procter2013reading}:

\begin{compactitem}
 \item \textbf{Support:} the author of the response supports the veracity of the rumour.
 \item \textbf{Deny:} the author of the response denies the veracity of the rumour.
 \item \textbf{Query:} the author of the response asks for additional evidence in relation to the veracity of the rumour.
 \item \textbf{Comment:} the author of the response makes their own comment without a clear contribution to assessing the veracity of the rumour.
\end{compactitem}

Prior work in the area has found the task difficult, compounded by the variety present in language use between different stories~\cite{lukasik2015classifying,zubiaga2017detection}.
This indicates it is challenging enough to make for an interesting SemEval shared task.

\begin{figure*}
\small
 \begin{framed}
  \textbf{Veracity prediction examples:} \vspace{1em}

\noindent \textbf{u1:} Hostage-taker in supermarket siege killed, reports say. \#ParisAttacks --LINK-- \textbf{[true]} \\

\noindent \textbf{u1:} OMG. \#Prince rumoured to be performing in Toronto today. Exciting! \textbf{[false]}
  
  \end{framed}
 \caption{Examples of source tweets with a veracity value, which has to be predicted in the veracity prediction task.}
 \label{fig:example-veracity}
\end{figure*}

\subsection{Subtask B - Veracity prediction}

The goal of this subtask is to predict the veracity of a given rumour. 
The rumour is presented as a tweet, reporting an update associated with a newsworthy event, but deemed unsubstantiated at the time of release. 
Given such a tweet/claim, and a set of other resources provided, systems should return a label describing the anticipated veracity of the rumour as true or false -- see Figure~\ref{fig:example-veracity}.

The ground truth of this task has been manually established by journalist members of the team who identified official statements or other trustworthy sources of evidence that resolved the veracity of the given rumour. 
Examples of tweets annotated for veracity are shown in Figure~\ref{fig:example-veracity}.

The participants in this subtask chose between two variants.
In the first case -- the \emph{closed} variant -- the veracity of a rumour had to be predicted solely from the tweet itself (for example \cite{liu2015real} rely only on the content of tweets to assess the veracity of tweets in real time, while systems such as Tweet-Cred \cite{Gupta:2014} follow a tweet level analysis for a similar task where the credibility of a tweet is predicted).
In the second case -- the \emph{open} variant -- additional context was provided as input to veracity prediction systems; this context consists of a Wikipedia dump.
Critically, no external resources could be used that contained information from after the rumour's resolution.
To control this, we specified precise versions of external information that participants could use.
This was important to make sure we introduced time sensitivity into the task of veracity prediction.
In a practical system, the classified conversation threads from Subtask~A could be used as context.

We take a simple approach to this task, using only true/false labels for rumours.
In practice, however, many claims are hard to verify; for example, there were many rumours concerning Vladimir Putin's activities in early 2015, many wholly unsubstantiable.
Therefore, we also expect systems to return a confidence value in the range of 0-1 for each rumour; if the rumour is unverifiable, a confidence of 0 should be returned.





\subsection{Impact}

Identifying the veracity of claims made on the web is an increasingly important task~\cite{zubiaga2015towards}.
Decision support, digital journalism and disaster response already rely on picking out such claims~\cite{procter2013reading}.
Additionally, web and social media are a more challenging environment than e.g. newswire, which has traditionally provided the mainstay of similar tasks (such as RTE~\cite{bentivogli2011seventh}). 
Last year we ran a workshop at WWW 2015, Rumors and Deception in Social Media: Detection, Tracking, and Visualization (RDSM 2015)\footnote{http://www.pheme.eu/events/rdsm2015/} which garnered interest from researchers coming from a variety of backgrounds, including natural language processing, web science and computational journalism.

\section{Data \& Resources}
To capture web claims and the community reaction around them, we take data from the ``model organism'' of social media, Twitter~\cite{tufekci2014big}.
Data for the task is available in the form of online discussion threads, each pertaining to a particular event and the rumours around it. 
These threads form a tree, where each tweet has a parent tweet it responds to.
Together these form a conversation, initiated by a source tweet (see Figure \ref{fig:example}). 
The data has already been annotated for veracity and SDQC following a published annotation scheme~\cite{zubiaga2016analysing}, as part of the {\sc Pheme} project~\cite{derczynski2014pheme}, in which the task organisers are partners.

\subsection{Training Data}
Our training dataset comprises 297 rumourous threads collected for 8 events in total, which include 297 source and 4,222 reply tweets, amounting to 4,519 tweets in total. These events include well-known breaking news such as the Charlie Hebdo shooting in Paris, the Ferguson unrest in the US, and the Germanwings plane crash in the French Alps.
The size of the dataset means it can be distributed without modifications, according to Twitter's current data usage policy, as JSON files.

This dataset is already publicly available~\cite{Zubiaga2016} and constitutes the training and development data.

\subsection{Test Data}

For the test data, we annotated 28 additional threads. These include 20 threads extracted from the same events as the training set, and 8 threads from two newly collected events: (1) a rumour that Hillary Clinton was diagnosed with pneumonia during the 2016 US election campaign, and (2) a rumour that Youtuber Marina Joyce had been kidnapped.

The test dataset includes, in total, 1,080 tweets, 28 of which are source tweets and 1,052 replies.
The distribution of labels in the training and test datasets is summarised in Table \ref{tab:stats}.

\begin{table}
 \small
 \centering
 \begin{tabular}{l l l l l}
 \multicolumn{5}{c}{\textbf{Subtask A}} \\
 \midrule
 & S & D & Q & C \\
 \midrule
 Train & 910 & 344 & 358 & 2,907 \\
 Test & 94 & 71 & 106 & 778 \\
 \bottomrule
 \end{tabular}
 \begin{tabular}{l l l l}
 \multicolumn{4}{c}{\textbf{Subtask B}} \\
 \midrule
 & T & F & U \\
 \midrule
 Train & 137 & 62 & 98 \\
 Test & 8 & 12 & 8 \\
 \bottomrule
 \end{tabular}
 \caption{Label distribution of training and test datasets.}
 \label{tab:stats}
\end{table}

\subsection{Context Data}

Along with the tweet threads, we also provided additional context that participants could make use of. The context we provided was two-fold: (1) \textbf{Wikipedia articles} associated with the event in question. We provided the last revision of the article prior to the source tweet being posted, and (2) \textbf{content of linked URLs}, using the Internet Archive to retrieve the latest revision prior to the link being tweeted, where available.

\subsection{Data Annotation}

The annotation of rumours and their subsequent interactions was performed in two steps. 
In the first step, we sampled a subset of likely rumourous tweets from all the tweets associated with the event in question, where we used the high number of retweets as an indication of a tweet being potentially rumourous. 
These sampled tweets were fed to an annotation tool, by means of which our expert journalist annotators members manually identified the ones that did indeed report unverified updates and were considered to be rumours. 
Whenever possible, they also annotated rumours that had ultimately been proven true or the ones that had been debunked as false stories; the rest were annotated as ``unverified''. 
In the second step, we collected conversations associated with those rumourous tweets, which included all replies succeeding a rumourous source tweet. The type of support (SDQC) expressed by each participant in the conversation was then annotated through crowdsourcing. The methodology for performing this crowdsourced annotation process has been previously assessed and validated \cite{zubiaga2015crowdsourcing}, and is further detailed in \cite{zubiaga2016analysing}. The overall inter-annotator agreement rate of 63.7\% showed the task to be challenging, and easier for source tweets (81.1\%) than for replying tweets (62.2\%).

The evaluation data was not available to those participating in any way in the task, and selection decisions were taken only by organisers not connected with any submission, to retain fairness across submissions.

Figure \ref{fig:example} shows an example of what a data instance looks like, where the source tweet in the tree presents a rumourous statement that is supported, denied, queried and commented on by others. 
Note that replies are nested, where some tweets reply directly to the source, while other tweets reply to earlier replies, e.g., u4 and u5 engage in a short conversation replying to each other in the second example. 
The input to the veracity prediction task is simpler than this; here participants had to determine if a rumour was true or false by only looking at the source tweet (see Figure \ref{fig:example-veracity}), and optionally making use of the additional context provided by the organisers.

To prepare the evaluation resources, we collected and sampled the tweets around which there is most interaction, placed these in an existing annotation tool to be annotated as rumour vs. non-rumour, categorised them into rumour sub-stories, and labelled them for veracity. 

For Subtask A, the extra annotation for support / deny / question / comment at the tweet level within the conversations were performed through crowdsourcing -- as performed to satisfactory quality already with the existing training data~\cite{zubiaga2015crowdsourcing}.


\section{Evaluation}

The two subtasks were evaluated as follows.

\paragraph{SDQC stance classification:} The evaluation of the SDQC needed careful consideration, as the distribution of the categories is clearly skewed towards comments. 
Evaluation is through classification accuracy. 

\paragraph{Veracity prediction:} The evaluation of the predicted veracity, which is either true or false for each instance, was done using macroaveraged accuracy, hence measuring the ratio of instances for which a correct prediction was made.
 Additionally, we calculated RMSE $\rho$ for the difference between system and reference confidence in correct examples and provided the mean of these scores.
 Incorrect examples have an RMSE of 1.
 This is normalised and combined with the macroaveraged accuracy to give a final score; e.g. $acc = (1-\rho) acc$.
 
The baseline is the most common class. For Task A, we also introduce a baseline excluding the common, low-impact ``comment'' class, considering accuracy over only support, deny and query. This is included as the SDQ baseline.
 
\section{Participant Systems and Results}

We have had 13 system submissions at RumourEval, eight submissions for Subtask~A \cite{kochkina2017turing,bahuleyan2017uwaterloo,srivastava2017dfki,wang2017ecnu,singh2017iitp,chen2017ikm,garcialozano2017mama,enayet2017niletmrg}, the identification of stance towards rumours, and five submissions for Subtask~B \cite{srivastava2017dfki,wang2017ecnu,singh2017iitp,chen2017ikm,enayet2017niletmrg}, the rumour veracity classification task, with participant teams coming from four continents (Europe: Germany, Sweden, UK; North America: Canada; Asia: China, India, Taiwan; Africa: Egypt), showing the global reach of the issue of rumour veracity on social media.

\begin{table}
\centering
\begin{tabular}{lr}
\hline
{\bf Team} & {\bf Score}\\
\hline
DFKI DKT &	0.635	 \\
ECNU &	0.778	 \\
IITP &	0.641	 \\
IKM	& 0.701	 \\
Mama Edha &	0.749	 \\
NileTMRG &	0.709	 \\
Turing &	0.784	 \\
UWaterloo &	0.780\\
\hline
Baseline (4-way) & 0.741 \\
Baseline (SDQ)   & 0.391 \\
\hline
\end{tabular}
\caption{Results for Task A: support/deny/query/comment classification.}
\label{tab:taskA}
\end{table}

Most participants tackled Subtask A, which involves classifying a tweet in a conversation thread as either supporting (S), denying (D), querying (Q) or commenting on (C) a rumour.
Results are given in Table~\ref{tab:taskA}
The distribution of SDQC labels in the training, development and test sets favours comments (see Table~\ref{tab:stats}.
Including and recognising the items that fit in this class is important for reducing noise in the other, information-bearing classifications (support, deny and query).
In actual fact, comments are often express implicit support; the absence of dispute is a soft signal of agreement.

Systems generally viewed this task as a four-way single tweet classification task, with the exception of the best performing system (Turing), which addressed it as a sequential classification problem, where the SDQC label of each tweet depends on the features and labels of the previous tweets, and the ECNU and IITP systems. 
The IITP system takes as input pairs of source and reply tweets whereas the ECNU system addressed class imbalance by decomposing the problem into a two step classification task (comment vs. non-comment), and all non-comment tweets classified as SDQ. 
Half of the systems employed ensemble classifiers, where classification was obtained through majority voting (ECNU, MamaEdha, UWaterloo, DFKI-DKT). 
In some cases the ensembles were hybrid, consisting both of machine learning classifiers and manually created rules, with differential weighting of classifiers for different class labels (ECNU, MamaEdha, DFKI-DKT). 
Three systems used deep learning, with team Turing employing LSTMs for sequential classification, team IKM using convolutional neural networks (CNN) for obtaining the representation of each tweet, assigned a probability for a class by a softmax classifier and team Mama Edha using CNN as one of the classifiers in their hybrid conglomeration. 
The remaining two systems NileTMRG and IITP used support vector machines with linear and polynomial kernel respectively. 
Half of the systems invested in elaborate feature engineering including cue words and expressions denoting Belief, Knowledge, Doubt and Denial (UWaterloo) as well as Tweet domain features including meta-data about users, hashtags and event specific keywords (ECNU, UWaterloo, IITP, NileTMRG). 
The systems with the least elaborate features were IKM and Mama Edha for CNNs (word embeddings), DFKI-DKT (sparse word vectors as input to logistic regression) and Turing (average word vectors, punctuation, similarity between word vectors in current tweet, source tweet and previous tweet, presence of negation, picture, URL).  
Five out of the eight systems used pre-trained word embeddings, mostly Google News word2vec embeddings, while ECNU used four different types of embeddings. Overall, elaborate feature engineering and a strategy for addressing class imbalance seemed to pay off, as can be seen by the success of the high performance of the UWaterloo and ECNU systems. The success of the best performing system (Turing) can be attributed both to the use of LSTM to address the problem as a sequential task and the choice of word embeddings.

\begin{table}
\centering
\begin{tabular}{lrr}
\hline
{\bf Team} & {\bf Score} & {\bf Confidence RMSE}\\
\hline
IITP	&0.393&	0.746 \\
\hline
\end{tabular}
\caption{Results for Task B: Rumour veracity - open variant.}
\label{tab:taskBopen}
\end{table}

\begin{table}
\centering
\begin{tabular}{lrr}
\hline
{\bf Team} & {\bf Score} & {\bf Confidence RMSE}\\
\hline
DFKI DKT&	0.393&	0.845\\
ECNU	&0.464	&0.736\\
IITP	&0.286	&0.807\\
IKM	&0.536	&0.763\\
NileTMRG	&0.536&	0.672\\
\hline
Baseline & 0.571 & -- \\
\hline
\end{tabular}
\caption{Results for Task B: Rumour veracity - closed variant.}
\label{tab:taskBclosed}
\end{table}

Subtask~B, veracity classification of a source tweet, was viewed as either a three-way (NileTMRG, ECNU, IITP) or two-way (IKM, DFKI-DKT) single tweet classification task.
Results are given in Table~\ref{tab:taskBopen} for the open variant, where external resources may be used,\footnote{Namely, the 20160901 English Wikipedia dump.} and Table~\ref{tab:taskBclosed} for the closed variant -- with no external resource use permitted. 
The systems used mostly similar features and classifiers to those in Subtask~A, though some added features more specific to the distribution of SDQC labels in replies to the source tweet (e.g. the best performing system in this task, NileTMRG, considered the percentage of reply tweets classified as either S, D or Q). 


\section{Conclusion}
Detecting and verifying rumours is a critical task and in the current media landscape, vital to populations so they can make decisions based on the truth.
This shared task brought together many approaches to fixing veracity in real media, working through community interactions and claims made on the web.
Many systems were able to achieve good results on unravelling the argument around various claims, finding out whether a discussion supports, denies, questions or comments on rumours.

The commentary around a story often helps determine how true that story is, so this advance is a great positive.
However, finding out accurately whether a story is false or true remains really tough.
Systems did not reach the most-common-class baseline, despite the data not being exceptionally skewed.
even the best systems could have the wrong level of confidence in a true/false judgment, weakly verifying stories that are true and so on.
This tells us that we are making progress, but that the problem is so far very hard.

RumourEval leaves behind competitive results, a large number of approaches to be dissected by future researchers, and a benchmark dataset of thousands of documents and novel news stories.
This sets a good baseline for the next steps in the area of fake news detection, as well as the material anyone needs to get started on the problem and evaluate and improve their systems.

\section*{Acknowledgments}
This work is supported by the European Commission's 7th Framework Programme for research, under grant No. 611223 {\sc Pheme}.
This work is also supported by the European Union’s Horizon 2020 research and innovation programme under grant agreement No. 687847 {\sc Comrades}.
We are grateful to Swissinfo.ch for their extended support in the form of journalistic advice, keeping the task well-grounded, and annotation and task design efforts.
We also extend our thanks to the SemEval organisers for their sustained hard work, and to our participants for bearing with us during the first shared task of this nature and all the joy and trouble that comes with it.

\bibliography{naaclhlt2016}

\begin{thebibliography}{}
\expandafter\ifx\csname natexlab\endcsname\relax\def\natexlab#1{#1}\fi

\bibitem[{Anand et~al.(2011)Anand, Walker, Abbott, Tree, Bowmani, and
  Minor}]{Anand:2011:CRD:2107653.2107654}
Pranav Anand, Marilyn Walker, Rob Abbott, Jean E.~Fox Tree, Robeson Bowmani,
  and Michael Minor. 2011.
\newblock \href{http://dl.acm.org/citation.cfm?id=2107653.2107654}{Cats rule
  and dogs drool!: Classifying stance in online debate}.
\newblock In {\em Proceedings of the 2Nd Workshop on Computational Approaches
  to Subjectivity and Sentiment Analysis\/}. Association for Computational
  Linguistics, Stroudsburg, PA, USA, WASSA '11, pages 1--9.
\newblock
  \href{http://dl.acm.org/citation.cfm?id=2107653.2107654}{http://dl.acm.org/citation.cfm?id=2107653.2107654}.

\bibitem[{Bahuleyan and Vechtomova(2017)}]{bahuleyan2017uwaterloo}
Hareesh Bahuleyan and Olga Vechtomova. 2017.
\newblock {UWaterloo at SemEval-2017 Task 8: Detecting Stance towards Rumours
  with Topic Independent Features}.
\newblock In {\em Proceedings of SemEval\/}. ACL.

\bibitem[{Bentivogli et~al.(2011)Bentivogli, Clark, Dagan, Dang, and
  Giampiccolo}]{bentivogli2011seventh}
Luisa Bentivogli, Peter Clark, Ido Dagan, Hoa Dang, and Danilo Giampiccolo.
  2011.
\newblock {The seventh Pascal Recognizing Textual Entailment challenge}.
\newblock In {\em Proceedings of the Text Analysis Conference\/}. NIST.

\bibitem[{Chen et~al.(2017)Chen, Liu, and Kao}]{chen2017ikm}
Yi-Chin Chen, Zhao-Yand Liu, and Hung-Yu Kao. 2017.
\newblock {IKM at SemEval-2017 Task 8: Convolutional Neural Networks for Stance
  Detection and Rumor Verification}.
\newblock In {\em Proceedings of SemEval\/}. ACL.

\bibitem[{Derczynski and Bontcheva(2014)}]{derczynski2014pheme}
Leon Derczynski and Kalina Bontcheva. 2014.
\newblock Pheme: Veracity in digital social networks.
\newblock In {\em UMAP Workshops\/}.

\bibitem[{Enayet and El-Beltagy(2017)}]{enayet2017niletmrg}
Omar Enayet and Samhaa~R. El-Beltagy. 2017.
\newblock {NileTMRG at SemEval-2017 Task 8: Determining Rumour and Veracity
  Support for Rumours on Twitter}.
\newblock In {\em Proceedings of SemEval\/}. ACL.

\bibitem[{Garc\'ia~Lozano et~al.(2017)Garc\'ia~Lozano, Lilja, Tj\"ornhammar,
  and Maja~Karasalo}]{garcialozano2017mama}
Marianela Garc\'ia~Lozano, Hanna Lilja, Edward Tj\"ornhammar, and Maja
  Maja~Karasalo. 2017.
\newblock {Mama Edha at SemEval-2017 Task 8: Stance Classification with CNN and
  Rules}.
\newblock In {\em Proceedings of SemEval\/}. ACL.

\bibitem[{Gupta et~al.(2014)Gupta, Kumaraguru, Castillo, and
  Meier}]{Gupta:2014}
Aditi Gupta, Ponnurangam Kumaraguru, Carlos Castillo, and Patrick Meier. 2014.
\newblock \href{https://doi.org/10.1007/978-3-319-13734-6\_16}{Tweetcred:
  Real-time credibility assessment of content on twitter}.
\newblock In {\em SocInfo\/}. pages 228--243.
\newblock
  \href{https://doi.org/10.1007/978-3-319-13734-6\_16}{https://doi.org/10.1007/978-3-319-13734-6\_16}.

\bibitem[{Hermida(2012)}]{hermida2012tweets}
Alfred Hermida. 2012.
\newblock Tweets and truth: Journalism as a discipline of collaborative
  verification.
\newblock {\em Journalism Practice\/} 6(5-6):659--668.

\bibitem[{Kochkina et~al.(2017)Kochkina, Liakata, and
  Augenstein}]{kochkina2017turing}
Elena Kochkina, Maria Liakata, and Isabelle Augenstein. 2017.
\newblock {Turing at SemEval-2017 Task 8: Sequential Approach to Rumour Stance
  Classification with Branch-LSTM}.
\newblock In {\em Proceedings of SemEval\/}. ACL.

\bibitem[{Kumar and Geethakumari(2014)}]{kumar2014detecting}
KP~Krishna Kumar and G~Geethakumari. 2014.
\newblock Detecting misinformation in online social networks using cognitive
  psychology.
\newblock {\em Human-centric Computing and Information Sciences\/} 4(1):1--22.

\bibitem[{Liu et~al.(2015)Liu, Nourbakhsh, Li, Fang, and Shah}]{liu2015real}
Xiaomo Liu, Armineh Nourbakhsh, Quanzhi Li, Rui Fang, and Sameena Shah. 2015.
\newblock Real-time rumor debunking on twitter.
\newblock In {\em Proceedings of the 24th ACM International on Conference on
  Information and Knowledge Management\/}. ACM, pages 1867--1870.

\bibitem[{Lukasik et~al.(2015)Lukasik, Cohn, and
  Bontcheva}]{lukasik2015classifying}
Michal Lukasik, Trevor Cohn, and Kalina Bontcheva. 2015.
\newblock Classifying tweet level judgements of rumours in social media.
\newblock In {\em Proceedings of the Conference on Empirical Methods in Natural
  Language Processing\/}. volume~2, pages 2590--2595.

\bibitem[{Mohammad et~al.(2016)Mohammad, Kiritchenko, Sobhani, Zhu, and
  Cherry}]{saif2016stance}
Saif~M Mohammad, Svetlana Kiritchenko, Parinaz Sobhani, Xiaodan Zhu, and Colin
  Cherry. 2016.
\newblock {SemEval-2016 Task 6: Detecting Stance in Tweets}.
\newblock In {\em Proceedings of the Workshop on Semantic Evaluation\/}.

\bibitem[{Moschitti et~al.(2015)Moschitti, Nakov, Marquez, Magdy, Glass, and
  Randeree}]{alessandromoschitti2015semeval}
Alessandro Moschitti, Preslav Nakov, Llu{\i}s Marquez, Walid Magdy, James
  Glass, and Bilal Randeree. 2015.
\newblock Semeval-2015 task 3: Answer selection in community question
  answering.
\newblock {\em SemEval-2015\/} page 269.

\bibitem[{Procter et~al.(2013{\natexlab{a}})Procter, Crump, Karstedt, Voss, and
  Cantijoch}]{procter2013police}
Rob Procter, Jeremy Crump, Susanne Karstedt, Alex Voss, and Marta Cantijoch.
  2013{\natexlab{a}}.
\newblock Reading the riots: What were the {Police doing on Twitter?}
\newblock {\em Policing and Society\/} 23(4):413--436.

\bibitem[{Procter et~al.(2013{\natexlab{b}})Procter, Vis, and
  Voss}]{procter2013reading}
Rob Procter, Farida Vis, and Alex Voss. 2013{\natexlab{b}}.
\newblock Reading the riots on twitter: methodological innovation for the
  analysis of big data.
\newblock {\em International journal of social research methodology\/}
  16(3):197--214.

\bibitem[{Qazvinian et~al.(2011)Qazvinian, Rosengren, Radev, and
  Mei}]{qazvinian2011rumor}
Vahed Qazvinian, Emily Rosengren, Dragomir~R Radev, and Qiaozhu Mei. 2011.
\newblock Rumor has it: Identifying misinformation in microblogs.
\newblock In {\em Proceedings of the Conference on Empirical Methods in Natural
  Language Processing\/}. Association for Computational Linguistics, pages
  1589--1599.

\bibitem[{Shao et~al.(2016)Shao, Ciampaglia, Flammini, and
  Menczer}]{shao2016hoaxy}
Chengcheng Shao, Giovanni~Luca Ciampaglia, Alessandro Flammini, and Filippo
  Menczer. 2016.
\newblock Hoaxy: A platform for tracking online misinformation.
\newblock {\em arXiv preprint arXiv:1603.01511\/} .

\bibitem[{Singh et~al.(2017)Singh, Narayan, Akhtar, Ekbal, and
  Bhattacharya}]{singh2017iitp}
Vikram Singh, Sunny Narayan, Md~Shad Akhtar, Asif Ekbal, and Pushpak
  Bhattacharya. 2017.
\newblock {IITP at SemEval-2017 Task 8: A Supervised Approach for Rumour
  Evaluation}.
\newblock In {\em Proceedings of SemEval\/}. ACL.

\bibitem[{Srivastava et~al.(2017)Srivastava, Rehm, and
  Moreno~Schneider}]{srivastava2017dfki}
Ankit Srivastava, Rehm Rehm, and Julian Moreno~Schneider. 2017.
\newblock {DFKI-DKT at SemEval-2017 Task 8: Rumour Detection and Classification
  using Cascading Heuristics}.
\newblock In {\em Proceedings of SemEval\/}. ACL.

\bibitem[{Tufekci(2014)}]{tufekci2014big}
Zeynep Tufekci. 2014.
\newblock Big questions for social media big data: Representativeness, validity
  and other methodological pitfalls.
\newblock In {\em Proceedings of the AAAI International Conference on Weblogs
  and Social Media\/}.

\bibitem[{Veil et~al.(2011)Veil, Buehner, and Palenchar}]{veil2011work}
Shari~R Veil, Tara Buehner, and Michael~J Palenchar. 2011.
\newblock A work-in-process literature review: Incorporating social media in
  risk and crisis communication.
\newblock {\em Journal of contingencies and crisis management\/}
  19(2):110--122.

\bibitem[{Wang et~al.(2017)Wang, Lan, and Wu}]{wang2017ecnu}
Feixiang Wang, Man Lan, and Yuanbin Wu. 2017.
\newblock {ECNU at SemEval-2017 Task 8: Rumour Evaluation Using Effective
  Features and Supervised Ensemble Models}.
\newblock In {\em Proceedings of SemEval\/}. ACL.

\bibitem[{Zhang et~al.(2015)Zhang, Zhang, Dong, Xiong, and
  Cheng}]{zhang2015automatic}
Qiao Zhang, Shuiyuan Zhang, Jian Dong, Jinhua Xiong, and Xueqi Cheng. 2015.
\newblock Automatic detection of rumor on social network.
\newblock In {\em Natural Language Processing and Chinese Computing\/},
  Springer, pages 113--122.

\bibitem[{Zubiaga et~al.(2017)Zubiaga, Aker, Bontcheva, Liakata, and
  Procter}]{zubiaga2017detection}
Arkaitz Zubiaga, Ahmet Aker, Kalina Bontcheva, Maria Liakata, and Rob Procter.
  2017.
\newblock Detection and resolution of rumours in social media: A survey.
\newblock {\em arXiv preprint arXiv:1704.00656\/} .

\bibitem[{Zubiaga et~al.(2015{\natexlab{a}})Zubiaga, Liakata, Procter,
  Bontcheva, and Tolmie}]{zubiaga2015crowdsourcing}
Arkaitz Zubiaga, Maria Liakata, Rob Procter, Kalina Bontcheva, and Peter
  Tolmie. 2015{\natexlab{a}}.
\newblock Crowdsourcing the annotation of rumourous conversations in social
  media.
\newblock In {\em Proceedings of the 24th International Conference on World
  Wide Web: Companion volume\/}. International World Wide Web Conferences
  Steering Committee, pages 347--353.

\bibitem[{Zubiaga et~al.(2015{\natexlab{b}})Zubiaga, Liakata, Procter,
  Bontcheva, and Tolmie}]{zubiaga2015towards}
Arkaitz Zubiaga, Maria Liakata, Rob Procter, Kalina Bontcheva, and Peter
  Tolmie. 2015{\natexlab{b}}.
\newblock Towards detecting rumours in social media.
\newblock In {\em Proceedings of the AAAI Workshop on AI for Cities\/}.

\bibitem[{Zubiaga et~al.(2016{\natexlab{a}})Zubiaga, Liakata, Procter, Hoi, and
  Tolmie}]{Zubiaga2016}
Arkaitz Zubiaga, Maria Liakata, Rob Procter, Geraldine Wong~Sak Hoi, and Peter
  Tolmie. 2016{\natexlab{a}}.
\newblock {PHEME rumour scheme dataset: Journalism use case}.
\newblock doi:10.6084/m9.figshare.2068650.v1.

\bibitem[{Zubiaga et~al.(2016{\natexlab{b}})Zubiaga, Liakata, Procter, Wong
  Sak~Hoi, and Tolmie}]{zubiaga2016analysing}
Arkaitz Zubiaga, Maria Liakata, Rob Procter, Geraldine Wong Sak~Hoi, and Peter
  Tolmie. 2016{\natexlab{b}}.
\newblock \href{https://doi.org/10.1371/journal.pone.0150989}{Analysing how
  people orient to and spread rumours in social media by looking at
  conversational threads}.
\newblock {\em PLoS ONE\/} 11(3):1--29.
\newblock
  \href{https://doi.org/10.1371/journal.pone.0150989}{https://doi.org/10.1371/journal.pone.0150989}.

\end{thebibliography}
\bibliographystyle{acl_natbib}
\end{document}